\documentclass[conference, 9pt]{IEEEtran}
\IEEEoverridecommandlockouts
\usepackage{cite}
\usepackage{amsmath,amssymb,amsfonts}
\usepackage{graphicx}
\usepackage{textcomp}
\usepackage{xcolor}
\usepackage[T1]{fontenc}
\usepackage{graphicx}
\usepackage{cite}
\usepackage{amsmath,amssymb,amsfonts}
\usepackage{textcomp}
\usepackage{xcolor}
\usepackage{subcaption}
\usepackage{url}
\usepackage{algpseudocode}

\def\BibTeX{{\rm B\kern-.05em{\sc i\kern-.025em b}\kern-.08em
    T\kern-.1667em\lower.7ex\hbox{E}\kern-.125emX}}
\begin{document}

% \title{Self-supervised conformal prediction for uncertainty quantification in
% Poisson-Gaussian imaging problems\\
% {\footnotesize \textsuperscript{*}Note: Sub-titles are not captured for https://ieeexplore.ieee.org  and

\title{Self-supervised conformal prediction for uncertainty quantification in
Poisson imaging problems
\thanks{This work was supported by French National Research Agency (ANR) (ANR-23-CE48-0009, ``\textit{OptiMoCSI}"); and by UKRI Engineering and Physical Sciences Research Council (EPSRC) (EP/V006134/1, EP/Z534481/1).}
}

\author{\IEEEauthorblockN{Bernardin Tamo Amougou}
\IEEEauthorblockA{\small\textit{School of Mathematical and Computer Sciences}\\
\textit{\& Maxwell Institute for Mathematical Sciences}\\
\textit{Heriot-Watt University}, Edinburgh, UK\\
\& Universit\'e de Paris Cit\'e, Paris, France  \\
bt2027@hw.ac.uk}
\and
\IEEEauthorblockN{Marcelo Pereyra}
\IEEEauthorblockA{\small\textit{School of Mathematical and Computer Sciences}\\
\textit{\& Maxwell Institute for Mathematical Sciences}\\
\textit{Heriot-Watt University}, Edinburgh, UK\\
 m.pereyra@hw.ac.uk}
\and
\IEEEauthorblockN{Barbara Pascal}
\IEEEauthorblockA{\small
	\textit{Nantes Université, École Centrale Nantes} \\
	\textit{CNRS, LS2N, UMR 6004}\\
    F-44000 Nantes, France \\
barbara.pascal@cnrs.fr}
}

\maketitle

\begin{abstract}
Image restoration problems are often ill-posed, leading to significant uncertainty in reconstructed images. Accurately quantifying this uncertainty is essential for the reliable interpretation of reconstructed images.  However, image restoration methods often lack uncertainty quantification capabilities. Conformal prediction offers a rigorous framework to augment image restoration methods with accurate uncertainty quantification estimates, but it typically requires abundant ground truth data for calibration. This paper presents a self-supervised conformal prediction method for Poisson imaging problems which leverages Poisson Unbiased Risk Estimator to eliminate the need for ground truth data. The resulting self-calibrating conformal prediction approach is applicable to any Poisson linear imaging problem that is ill-conditioned, and is particularly effective when combined with modern self-supervised image restoration techniques trained directly on measurement data. The proposed method is demonstrated through numerical experiments on image denoising and deblurring; its performance are comparable to supervised conformal prediction methods relying on ground truth data.
\end{abstract}


%

%
%\titlerunning{SURE-based Conformal Prediction for UQ in Imaging}
%\author{Jasper M. Everink\inst{1}\orcidID{0000-0001-7263-0317} \and Bernardin Tamo Amougou\inst{2,3} \and Marcelo Pereyra\inst{2}\orcidID{0000-0001-6438-6772} }
%
%\authorrunning{J.M. Everink, B. Tamo Amougou and  M. Pereyra.}
%\institute{Technical University of Denmark, Kgs. Lyngby, Denmark, \email{jmev@dtu.dk} \and Heriot-Watt University \& Maxwell Institute for Mathematical Sciences, Edinburgh, UK \and Universit\'e de Paris Cit\'e, Paris, France }

%
\maketitle              % typeset the header of the contribution

\section{Introduction}
Poisson imaging problems are widely encountered across various scientific and engineering disciplines, with notable prevalence in fields such as astronomy and microscopy \cite{Starck2006-ew, Bertero2009}; common examples include Poisson image denoising, deblurring and computed tomography \cite{Figueiredo2010, khademi2021self, savanier2023deep}. Such problems are often not well posed, meaning that there exists a wide range of possible solutions that are in agreement with the observed data. Accounting for this inherent uncertainty is essential when using the reconstructed images as evidence for science  or critical decision-making \cite{Robert2007}. Unfortunately, most methods for solving Poisson imaging problems currently available are unable to accurately quantify the uncertainty in the delivered solutions.
%, from image deblurring~\cite{khademi2021self, laroche2023deep} to computed tomography~\cite{savanier2023deep}. In such problems, there exists a large set of solutions, possibly significantly different, which are equally probable given a corrupted observation.
%When restored images are used to guide sensitive decisions, accounting for this uncertainty is crucial and has triggered massive research efforts in imaging sciences.

Bayesian statistical inference strategies, which rely on prior knowledge about the likely solutions to the problem as encoded by a probability distribution, are the predominant approach to uncertainty quantification in imaging problems (see, e.g., \cite{Figueiredo2010, Marnissi2017,melidonis2023efficient,melidonis2025scorebased}). Bayesian imaging techniques can deliver a wide range of inferences, and the most accurate solutions to Poisson imaging problems are currently obtained by using Bayesian imaging methods combining ideas from optimisation, deep learning and stochastic sampling \cite{melidonis2025scorebased}. However, even the most advanced Bayesian imaging strategies currently available struggle to provide accurate uncertainty quantification on structures larger than a few pixels in size~\cite{thong2024bayesianimagingmethodsreport}. This has stimulated research on other frameworks for performing uncertainty quantification in imaging inverse problems. Namely, bootstrapping and conformal prediction have recently emerged as promising frameworks for delivering accurate uncertainty quantification for large image structures.%, in particular confidence regions, that are well calibrated and robust to experimental replication.

With regards to bootstrapping, we note the recent equivariant bootstrapping method~\cite{tachella2023equivariant} which yields excellent confidence regions, even for large image structures.
Equivariant bootstrapping is a statistical resampling strategy that takes advantage of known symmetries of the problem to significantly reduce the bias inherent to synthetic resampling. It does not require ground truth data for calibration, and it can be therefore applied to many quantitative and scientific imaging problems for which obtaining reliable ground truth data is difficult or impossible. Equivariant bootstrapping is effective in problems that are severely ill-posed, for example in compressive sensing, inpainting, limited angle tomography or radio-interferometry \cite{tachella2023equivariant,Liaudat2024}. Conversely, it often performs poorly for problems which are potentially severely ill-conditioned but not ill-posed, as this prevents leveraging symmetries to remove the bootstrapping bias ~\cite{tachella2023equivariant}.

In contrast, conformal prediction is a fully data-driven strategy for constructing regions with almost exact marginal coverage~\cite{angelopoulos2021gentle}. Unlike equivariant bootstrapping, conformal prediction is agnostic to both the image observation model and our prior knowledge about likely solutions. However, in its original form, conformal prediction requires abundant ground truth data for calibration, which is a main bottleneck to implementation for quantitative and scientific imaging problems where ground truth data is often scarce or unavailable. The need for ground truth can be partially mitigated by combining conformal prediction with Bayesian inference and by focusing on pixels of a single image rather than regions, as proposed in \cite{Narnhofer2024}. Alternatively, the recent method~\cite{everink2025} fully bypasses the need for ground truth data by focusing on  Gaussian imaging problems and leveraging  Stein's Unbiased Risk Estimator~\cite{9054593} to perform conformal prediction directly from the observed measurement data, in a self-supervised manner. The experiments reported in \cite{everink2025} show that, for non-blind linear Gaussian imaging problems, self-supervised conformal prediction is as accurate as supervised conformal prediction. 

There are many important imaging problems for which the Gaussian noise assumption underpinning \cite{everink2025} is not sufficiently accurate to deliver meaningful inferences. In particular, imaging sensors often exhibit so-called \emph{shot} noise, a form of signal-dependent measurement error that exhibits Poisson statistics. With quantitive and scientific Poisson imaging problems in mind, this paper proposes a generalisation of the self-supervised conformal prediction framework \cite{everink2025} for high-dimensional linear inverse problems involving Poisson noise. 

The remainder of the paper is organised as follows. The principles of conformal prediction for inverse problems involving Poisson noise are introduced in Section~\ref{sec:background}. Then, the proposed self-supervised conformal prediction framework is detailed in Section~\ref{sec:method}. 
This framework is demonstrated on Poisson image denoising and deblurring experiments in Section~\ref{sec:experiments}.
Concluding remarks and future research directions are provided in Section~\ref{sec:discussion}.

\clearpage
\newpage
\section{Conformal Prediction under Poisson models}
\label{sec:background}
In the context of image estimation, conformal prediction aims to identify a set of statistically likely values for an unknown image of interest $x^{\star}$ from some noisy measurements $y \in \mathbb{R}^m$ of $x^{\star}$. Assuming $x^{\star}$ and $y$ are realizations of random variables $X$ and $Y$, for a confidence level \( \alpha \in (0,1) \), we seek a region $C(Y) \subset \mathbb{R}^n$ verifying
\begin{equation}\label{predictionC}
    \mathbb{P}_{(X, Y)}\left(X \in C(Y) \right) \geq 1-\alpha\,
\end{equation}
where the probability is taken w.r.t. the joint distribution of \( (X, Y) \).
Conformal prediction is a general framework leveraging a \emph{non-conformity measure} \( s: \mathbb{R}^n \times \mathbb{R}^m \to \mathbb{R} \), which reflects how plausible is an image $x$ given measurements $y$, to provide such confidence sets~\cite{angelopoulos2021gentle}.
Once chosen $s$, the set \( C(y) \) is constructed by computing the top \((1-\alpha)\)-quantile \( q_\alpha \) of the statistic \( s(X, Y) \), i.e.,
\begin{align}
\label{eq:Cy}
C(y) := \{x \in \mathbb{R}^n \,|\, s(x, y) \leq q_\alpha\} \quad \text{for all } y \in \mathbb{R}^m.
\end{align}
which by construction set satisfies:
\begin{align}
\label{eq:proba}
\mathbb{P}_{(X, Y)} \big(X \in C(Y)\big) = \mathbb{P}_{(X, Y)} \big(s(X, Y) \leq q_\alpha \big) \geq 1 - \alpha,
\end{align}
for any confidence level \( \alpha \in (0, 1) \). 
Provided that $s$ has been suitably designed and that enough samples \(\{x_i, y_i\}_{i=1}^M\) under the joint distribution of $(X,Y)$ are available to calibrate $q_\alpha$, then~\eqref{eq:Cy} yields regions $C(y)$ that contain $x^\star$ with the prescribed probability.

In this paper, we consider a linear Poisson, i.e., \emph{shot} noise, observation model, in which $X$ and $Y$ are related by
\[
(Y \mid X = x^{\star}) \sim \mathcal{P}\!\Bigl(\gamma A x^{\star}\Bigr)\, ,
\]
where \(\mathcal{P}\) denotes the Poisson distribution, \(A\) is a linear measurement operator, and \(\gamma > 0\) is related to the strength of the shot noise (the larger $\gamma$, the easier the image restoration problem). We henceforth assume $A$ and $\gamma$ are known, and that $A$ is full rank but possibly severely ill-conditioned. 
To design $C(y)$, we assume the availability of some estimator of $x^{\star}$, henceforth denoted $\widehat{x}(y)$.

To gain an intuition for \eqref{predictionC}, it is useful to consider a specific example. 
Suppose that $x^{\star}$ is a high-resolution Positron Emission Tomography (PET) scan of an adult brain which is only accessible in practice through a noisy degraded measurement $y$.
Measurements are processed by the estimator \( \widehat{x}(y) \) to yield an estimate of \( x^\star \).
In this setting, $x^\star$ can be viewed as a realization of the random variable $X$ describing the distribution of generic noise-free and resolution-perfect brain PET scans of adult individuals within a studied population.
Though, rather than a point estimate, for diagnosis purpose one might be preferably interested in all plausible estimates of $x^{\star}$, by definition constituting the set $C(y)$ of Equation~\eqref{eq:Cy}.
Then, if a large number of brain PET scans are considered, at least a proportion $1-\alpha$ of the true images $x^{\star}$ should fall in the plausible sets $C(y)$.

% For instance, suppose that \( x^\star \) is a high-resolution Positron Emission Tomography (PET) scan of an adult brain, then it is a generic sample under the distribution of noise-free and resolution-perfect brain PET scans within the studied population.
% Then, the random variable \( X \) characterises the distribution of brain PET scans for a generic individual within the population, as obtained by an ideal noise-free and resolution-perfect PET scanner. 
%The specific image \( x^\star \) corresponds to a PET scan of a particular individual, while \( y \) represents the noisy, degraded measurement acquired in practice. 
% The estimator \( \widehat{x}(y) \) produces an estimate of \( x^\star \). 
%The region \( C(y) \) encapsulates a set of likely solutions, rather than a single estimate, and satisfies the guarantee in \eqref{predictionC}. This means that if the procedure is repeated across a large number of individuals from the population, the constructed regions \( C \) will contain the respective true images \( x^\star \) in at least \( 1-\alpha\) of the cases.

Split conformalisation is the most widely used implementation of conformal prediction. It consists in estimating the top \(\frac{\lceil(M+1)(1-\alpha)\rceil}{M}\)-quantile \(\hat{Q}_\alpha\) of \(\{s(x_i, y_i)\}_{i=1}^M\) from $M$ training samples \(\{x_i, y_i\}_{i=1}^M\) of independent (or exchangeable) realizations of \((X, Y)\).
Then, given a new measurement $Y_{\text{new}}$ stemming from some unknown image $X_{\text{new}}$, the set yielded by split conformal prediction is given by
\[
\widehat{C}(Y_{\text{new}}) := \{X_{\text{new}} \in \mathbb{R}^n \,|\, s(X_{\text{new}}, Y_{\text{new}}) \leq \hat{Q}_\alpha\},
\]
which satisfies the condition:
\begin{equation}\label{eq:conformal_guarantee}
    \mathbb{P}_{(X, Y)^{M+1}}\left(X_{\text{new}} \in \widehat{C}(Y_{\text{new}}) \right)  
    \geq 1-\alpha\,,
\end{equation}
where $(X, Y)^{M+1}$ denotes the joint distribution of the \( M \) training samples and the new observation.
We refer the reader to~\cite{angelopoulos2021gentle} for a comprehensive introduction to the conformal prediction framework.

One major advantage of the conformal prediction framework lies in its flexibility, notably in choosing the non-conformity measure.
Different choices for $s$ yield infinitely many different regions $\widehat{C}(y)$ satisfying~\eqref{eq:conformal_guarantee}.
Though, in practice, not all these plausible sets are useful neither relevant.
Indeed, as $\widehat{C}(y)$ consists in a sub-level set of the function $x\mapsto s(x,y)$ in high-dimension, many of the plausible sets might be overly large, up to encompassing most of the support of X, and hence poorly informative to practitioners.
Hence a careful design of $s$ is key to obtain compact, thus useful, sets. In particular, $s$ should be constructed so as to minimize the variability of $s(X,Y)$.
To that aim, normalized non-conformity measures~\cite{johnstone2021conformal} of the form
\begin{equation}\label{eq:score_multi_target}
s(x, y) = \|x - \widehat{x}(y)\|^2_{\Sigma(y)} = \left(x - \widehat{x}(y)\right)^\top \Sigma(y) \left(x - \widehat{x}(y)\right),
\end{equation}
where \( \Sigma(y) \) is a positive definite matrix of size \( n \times n \) are particularly useful, notably when \( \Sigma(y) \) enables significant reduction of variability in $s(X,Y)$.
An efficient strategy consists in choosing \( \Sigma(y) \) as the inverse covariance matrix of the estimation error $X - \widehat{x}(Y)$~\cite{johnstone2021conformal}.

A major practical bottleneck in the implementation of conformal prediction strategy is the difficulty, if not the impossibility, to get an abundance of reliable data \(\{s(x_i, y_i)\}_{i=1}^M\); indeed, obtaining the ground truth $x_i$ from measurements $y_i$ is precisely the overarching goal of an imaging problem. Also, relying on a training dataset might result in inaccurate prediction sets when a distribution shift occurs. In the example of brain PET scans, this corresponds to a situation in which the calibration dataset has been done on a population that represents poorly the population encountered during deployment. 

To bypass the need for ground-truth data, Everink et al. \cite{everink2025} recently proposed a self-supervised conformal prediction method for linear imaging problems with Gaussian noise. This paper generalises this approach to perform conformalised uncertainty quantification in linear Poisson imaging problems without ground truth data available.

\section{Self-supervised uncertainty quantification}\label{sec:method}
In a manner akin to \cite{everink2025}, we propose a self-supervised conformal prediction strategy that removes the need for ground-truth data by making use of a statistical estimator of \eqref{eq:score_multi_target}. Namely, we leverage a Poisson Unbiased Risk Estimator (PURE) \cite{LUISIER2010415}, which provides accurate and unbiased estimates of \eqref{eq:score_multi_target}. We propose to apply this estimator individually to a collection of \(M\) statistically exchangeable imaging problems of the form $(Y \mid X = x^{\star}_i) \sim  \,\mathcal{P}\!({\gamma A x^{\star}_i})$. 

To define the non-conformity measure \eqref{eq:score_multi_target}, we set \(\Sigma(y) = A^\top A\) to approximate the error inverse-covariance, as we expect estimates of $x^{\star}$ to be most accurate in the dominant eigenvector directions of \(A^\top A\) while the estimation error is likely to concentrate along its weaker eigenvectors. The resulting score is given by:
\begin{equation}\label{score}
s(x, y) = \frac{1}{m}\,\|A x - A \widehat{x}(y)\|_2^2,
\end{equation}
where we recall that \(m = \textrm{dim}(y)\) and that \(\widehat{x}(y)\) is an estimator of $x^{\star}$. Note that \(A\) being full rank is crucial, as otherwise, \(\Sigma(y)\) would be only positive semidefinite leading to unbounded prediction sets.

% Our proposed self-supervised conformal prediction method circumvents the need for ground truth data by leveraging Poisson unbiased risk estimate (PURE) \cite{LUISIER2010415}. 

% We begin by pooling together $M$ exchangeable imaging problems, where each problem involves an unknown image $x^{\star}_i$ and an observation $y_i$ which we consider to be a realisation of the conditional random variable $(Y|X=x^{\star}_i) \sim \gamma \mathcal{P}\left(\frac{Ax^{\star}_i}{\gamma}\right)$. To specify the non-conformity measure, we  consider $\Sigma(y)=A^\top A$ which is a natural choice for approximation for the error inverse-covariance when $(Y|X=x^{\star}_i) \sim \gamma \mathcal{P}\left(\frac{Ax^{\star}_i}{\gamma}\right)$, as we expect $\widehat{x}(Y)$ to be accurate along the leading eigenvectors of $A^\top A$ and the estimation error to concentrate along weak eigenvectors of $A^\top A$. This leads to the non-conformity measure
% \begin{equation} \label{score}
%   s(x, y) = \frac{1}{m}\|Ax - A\widehat{x}(y)\|_2^2,  
% \end{equation}
% where we recall that $A$ is assumed full-rank, but potentially very poorly conditioned . We require $A$ to be full rank as otherwise $\Sigma(y)$ is only positive semidefinite, implying that the corresponding prediction set can be unbounded.

To perform calibration without relying on ground-truth data, we replace the direct sample quantiles of \(\{s(x_i, y_i)\}_{i=1}^M\) with PURE-based estimates derived solely from the observed measurements \(\{y_i\}_{i=1}^M\).  Assuming that the estimator \(\widehat{x}\) is differentiable almost everywhere, and let $\partial\bigl(A\widehat{x}(y)\bigr)/\partial y$ denote the diagonal of the Jacobian of $y\mapsto A\widehat{x}(y)$, PURE for \eqref{score} takes the form
\begin{equation}\label{eq:SURE}
    \mathrm{PURE}(y) 
    = \frac{1}{m}\,\left\lVert \frac{y}{\gamma} - A \widehat{x}(y)\right\rVert_2^2 
    -\frac{1}{\gamma^2 m}\,\mathbf{1}^\top y
    +\frac{2 }{ \gamma m}y^\top\,\frac{\partial}{\partial y}\bigl(A\widehat{x}(y)\bigr).
\end{equation}
Following a law of large numbers argument, in settings with large \(m = \mathrm{dim}(y)\), PURE not only remains unbiased but also exhibits a low variance (this is analysed analytically for the Gaussian case in \cite{stein1981estimation} and verified empirically in Poisson problems \cite{Montagner2014}). Consequently, we expect that calibrating conformal prediction by using PURE quantiles will yield prediction sets in close agreement with the true quantiles of \(s(X, Y)\), thereby preserving the intended coverage properties. Although this introduces a small bias in the resulting quantiles, the effect is minor (the unbiased errors from PURE act as mild \emph{smoothing} on the sample quantiles). 

Adopting a split-conformal strategy, we compute \(\widehat{x}(y_i)\) and \(\mathrm{PURE}(y_i)\) for \(i \in \{1,\dots,M\}\). Then, for any $\alpha \in (0,1)$, we set
\[
    \widehat{C}(y_i) 
    \;=\; \{\,x \in \mathbb{R}^n : \|A x - A \widehat{x}(y_i)\|_2^2 / m \;\le\; \hat{Q}_\alpha^{(i)}\},
\]
where \(\hat{Q}_\alpha^{(i)}\) is the \(\tfrac{\lceil (M+1)(1-\alpha)\rceil}{M}\)-quantile of \(\{\mathrm{PURE}(y_j)\}_{j=1}^M\) with the \(i\)th term removed. Although PURE introduces some estimation bias, it is typically minor compared to likely distribution shifts in real deployment. Furthermore, the required \(\mathrm{PURE}(y_i)\) values can be computed in parallel. The full algorithm is summarized in Fig.~\ref{alg:one}.

For most estimators $\widehat{x}(Y)$ of practical interest, computing the PURE estimate typically involves a numerical approximation of $\frac{\partial}{\partial y}\bigl(A\widehat{x}(y)\bigr)$. Monte Carlo PURE (MC-PURE) \cite{chen2022robustequivariantimagingfully} is perhaps the most widely used strategy to implement PURE, but we find that it requires careful hyper-parameter tuning. Instead, we recommend adopting a more robust alternative based on Hutchinson’s stochastic trace approximation \cite{9054593}, which in the considered setting yields 
\begin{align}
  \frac{1}{m}  y^\top \frac{\partial}{\partial y}\bigl(A\widehat{x}(y)\bigr)
    \;=\;& \mathbb{E}_{\tilde{n} \sim \mathcal{N}\left(\boldsymbol{0}, \operatorname{diag}(y)\right)} \Bigl[\tilde{n}^\top \boldsymbol{J}_{h(y)} \,\tilde{n}\Bigr] \\
    \approx\;& \frac{1}{m} \sum_{i=1}^{m} \tilde{n}_i^\top \boldsymbol{J}_{h(y)}\,\tilde{n}_i
    \;\;\;\;\;\text{(Hutchinson’s estimator)} \\
    \approx\;& \frac{1}{m}\,\tilde{n}^\top \boldsymbol{J}_{\tilde{n}^\top h(y)}, \label{eq:MC-SURE_N}
 \end{align}
where the last step is efficiently computed via automatic differentiation \cite{9054593}. In our experience, this approach provides accurate and computationally efficient estimates, even in large-scale settings.

\begin{figure}
\begin{algorithmic}[1]
\Require{Forward operator $A$, noise variance $\sigma^2$, estimator $\widehat{x}$, measurement $y$, samples $\{y_i\}_{i = 1}^M$, precision level $1-\alpha \in (0,1) . $}  
\Statex
\For{$i \gets 1$ to $M$}                    
    \State {$S_i \gets \textrm{PURE}(y_i)$ using \eqref{eq:SURE} and \eqref{eq:MC-SURE_N}}
    % \If{$s_{\text{mul}}$}
    %     \State $s_i \gets \frac{\|Ax_i-A\widehat{x}(y_i)\|_2^2}{S_i}$
    % \ElsIf{$s_{\text{add}}$}
    %     \State $s_i \gets \|Ax_i-A\widehat{x}(y_i)\|_2^2 - S_i$
    % \EndIf
\EndFor

\State {$\hat{Q}_\alpha$ $\gets$ top $\frac{\lceil(M+1)(1-\alpha)\rceil}{M}$-quantile of $\{S_i\}_{i = 1}^M$}
\State $m  \gets dim(y)$

% \State {$S \gets \textrm{SURE}(y)$ using \eqref{eq:SURE} and \eqref{eq:MC-SURE_2}}

% \If{$s_{\text{mul}}$}
\State {$\widehat{C}(y)$ $\gets$ $\{x\in \mathbb{R}^n\,|\, \|Ax-A\widehat{x}(y)\|_2^2 / m\leq  \hat{Q}_\alpha\}$}
%% \ElsIf{$s_{\text{add}}$}
% \State {$\widehat{C}(y)$ $\gets$ $\{x\in \mathbb{R}^n\,|\, \|Ax-A\widehat{x}(y_{\text{new}})\|_2^2 \leq S + \hat{Q}_\alpha\}$}
% \EndIf
\Statex 
\Ensure{Prediction set $\widehat{C}(y)$}
\end{algorithmic}
\caption{\label{alg:one}PURE-based conformal prediction}
\end{figure}

%\newpage
\section{Poisson imaging  experiments}\label{sec:experiments}
We illustrate our proposed self-supervised conformal prediction framework on two canonical Poisson image image restoration tasks: Poisson image denoising and non-blind Poisson image deblurring. In each case, we employ the algorithm of Fig.~\ref{alg:one} to construct conformal prediction sets over a range of confidence levels \(\alpha \in (0, 1)\), sampled from \(0\%\) to \(100\%\). By design, each prediction set should contain the ground-truth solution with a probability close to \(1 - \alpha\). To verify this in practice, we compute empirical coverage probabilities on a separate test set. Specifically, for each \(\alpha\), we measure the proportion of images in the test set that lie within their corresponding prediction sets. In both experiments, the calibration sets are of size \(M = 900\), which is sufficiently large to ensure that comparisons with supervised methods are fair; the test set is of size \(M = 200\). (We chose this split between calibration and test set to facilitate comparisons with supervised alternatives; however, our self-supervised method can also be naturally implemented by using a leave-one-out strategy). With regards to the choice of $\widehat{x}$, for both experiments we use a deep neural network estimator trained end-to-end in a self-supervised manner by using the PyTorch Deep Inverse library\footnote{\url{https://deepinv.github.io/deepinv/}}. More precisely, we train $\widehat{x}$ by using the PURE-based self-supervised training approach described in~\cite{chen2022robustequivariantimagingfully}. As a baseline for comparison, we also report results using a fully supervised conformal prediction technique implemented by using the same estimator $\widehat{x}$ but trained in a supervised manner, so where ground-truth data are available both for training the estimator as well as for calibrating the conformal prediction sets.

\begin{figure}[t!]
    \centering
    \includegraphics[width=\linewidth]{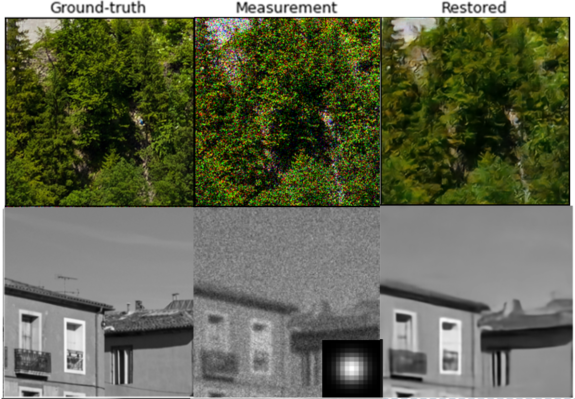}
    \caption{{Image reconstruction results for image denoising (top) and non-blind image deblurring (bottom), by using a self-supervised neural-network estimator $\hat{x}(y)$ and images from \texttt{DIV2K}.}}
    \label{fig:reconstruction}
\end{figure}

%, and the results highlight the ability of the proposed method to provide accurate uncertainty quantification across a range of restoration tasks.

% We now illustrate the proposed conformal prediction approach by applying it to three image restoration problems: image denoising, image deblurring and computed tomography. To show the versatility of the method, we consider the construction of conformal prediction sets for both model-driven and data-driven image restoration techniques. We conduct the following image restoration experiments using the deep inverse library \cite{Tachella_DeepInverse_A_deep_2023} :

\subsection{Image Poisson  Denoising}
We consider colour images of size $d=256\times 256$ pixels taking values in the hypercube $[0,1]^{3d}$, obtained by cropping images from the \texttt{DIV2K} dataset \cite{Agustsson_2017_CVPR_Workshops}, which we artificially corrupt  by applying the considered Poisson observation model with $A = \mathbb{I}$ and \( \gamma = 4 \). As image restoration method, we use an unrolled proximal gradient network with $4$ iterations and no weight-tying across iterations with a backbone denoiser  set as the U-Net architecture with 2 scales. The training is done  in a self-supervised manner by using the PURE loss \cite{chen2022robustequivariantimagingfully}. The training data consists of $900$ noisy measurements, we do not use any form of ground truth data neither for training nor for conformal prediction calibration. Fig. \ref{fig:reconstruction} (top) shows an example of a clean image, its noisy measurement, and the estimated reconstruction. 

We then use these same noisy measurements to compute our proposed self-supervised conformal prediction sets, and assess their accuracy empirically by using $200$ measurement-truth pairs from the test dataset. The results are reported in Fig. \ref{fig:denoising} below, together with the results obtained by using the equivalent supervised conformal prediction approach that relies on ground truth data for calibration and the supervised version of the estimator. We observe that our method delivers prediction sets that are remarkably accurate and in close agreement with the results obtained by using supervised conformal prediction, demonstrating that the bias stemming from using PURE instead of ground truth data is negligible.%, at least in this case. %For completeness, Fig. \ref{fig:histogram} (top) shows the empirical distribution of the non-similarity function $s(x,y)$ for the supervised conformal prediction based on the MSE, and the proposed self-supervised conformal prediction based on a PURE estimate of the MSE. Again, we observe close agreement between these distributions.%, with the PURE distribution being slightly more spread due to the random error inherent to PURE. 

%Moreover, for completeness we also report the results obtained by equivariant bootstrapping \cite{tachella2023equivariant}, which is significantly less accurate for this problem\footnote{The bootstrap is implemented by using rotation and two-dimensional shifts. Rotations are sampled from a Gaussian distribution with zero mean and a standard deviation of \( \sigma_{\theta} \) (denoising: \( \sigma_{\theta} = 125 \), deblurring: \( \sigma_{\theta} = 200 \)), while horizontal and vertical shifts are sampled from a uniform distribution on \( [-\Delta t, \Delta t] \) pixels (denoising: \( \Delta t = 250 \)}. The lack of accuracy of equivariant bootstrapping stems 

\begin{figure}[t!]
    \centering
    \includegraphics[width=1.1\linewidth]{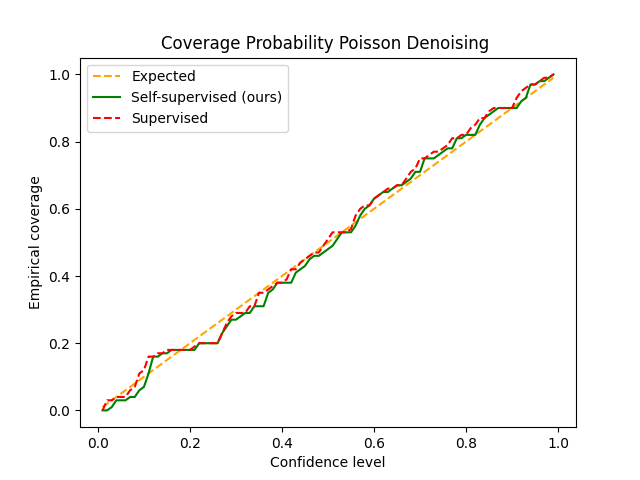}
    \caption{\textbf{Poisson image denoising experiment}: desired confidence level vs empirical coverage; the proposed self-supervised conformal prediction methods deliver prediction sets with near perfect coverage.}
    \label{fig:denoising}
    \vspace{-3mm}
\end{figure}

\subsection{Image Poisson Deblurring}
We now consider a non-blind Poisson image deblurring experiment with grayscale  images of size $256 \times 256$ pixels taking values in the hypercube $[0,1]^{d}$, also derived from the \texttt{DIV2K} dataset \cite{Agustsson_2017_CVPR_Workshops} and artificially degraded with an isotropic Gaussian blur of standard deviation  $\sigma = 2$ pixels, and Poisson noise  with \( \gamma = 60 \). As our reconstruction network, we use the state-of-the-art SwinIR architecture \cite{SwinIR}, trained with the loss function proposed \cite{scanvic2024selfsupervisedlearningimagesuperresolution}. This loss leverages both scale-invariance and an unbiased risk estimator (PURE) to form a self-supervised training objective. Fig. \ref{fig:reconstruction} (bottom) depicts an example of a clean image, its noisy measurement, and its estimate $\hat{x}(y)$. 

\begin{figure}[t!]
    \centering
    \includegraphics[width=1.1\linewidth]{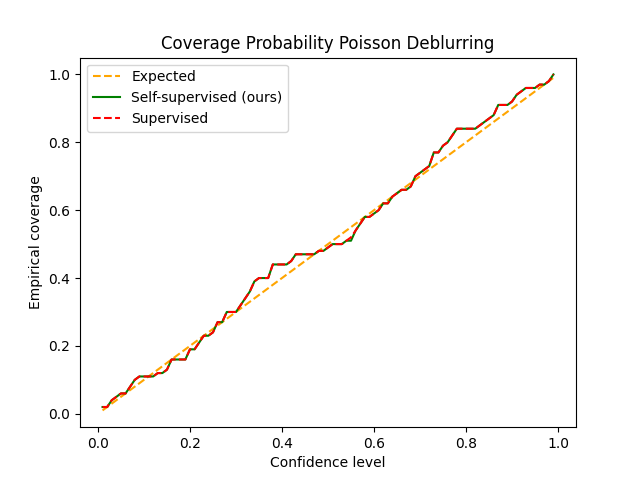}
    \caption{\textbf{Poisson image deblurring experiment}: desired confidence level vs empirical coverage; the proposed self-supervised conformal prediction methods deliver prediction sets with near perfect coverage.}
    \label{fig:deblurring}
    \vspace{-3mm}
\end{figure}

Again, we use 900 blurred and noisy images for training $y \mapsto \hat{x}(y)$ and for computing our proposed self-supervised conformal prediction sets. To evaluate their accuracy, we use 200 measurement–truth pairs from the test dataset. The corresponding results are presented in Figure~\ref{fig:deblurring}, alongside those obtained using an analogous supervised conformal prediction approach that relies on ground truth data for training and calibration. As in the previous experiment, our method produces prediction sets that closely match those from the fully supervised approach, indicating that the bias introduced by using PURE in place of ground truth data is again negligible.

%For completeness, Figure~\ref{fig:histogram} (down) displays the empirical distribution of the non-similarity function \(s(x, y)\) for both the supervised conformal prediction (based on MSE) and our self-supervised conformal prediction (based on a PURE estimate of MSE). The close agreement between these two distributions further demonstrates the effectiveness of our self-supervised approach.

% We use $900$ blurred and noisy images for the training as well as  to compute our proposed self-supervised conformal prediction sets, and assess their accuracy empirically by using $200$ measurement-truth pairs from the test dataset. The results are reported in Fig. \ref{fig:deblurring} below, together with the results obtained by using the equivalent supervised conformal prediction approach that relies on ground truth data for calibration. Again, we observe that our method delivers prediction sets that are accurate and remarkably close to the results obtained by using supervised conformal prediction with the supervised model, demonstrating that the bias stemming from using PURE instead of ground truth data is again negligible in this case. For completeness, Fig. \ref{fig:histogram} (right) below shows the empirical distribution of the non-similarity function $s(x,y)$ for the supervised conformal prediction based on the MSE, and the proposed self-supervised conformal prediction based on a PURE estimate of the MSE. Again, we observe close agreement between these distributions.

\section{Discussion and Conclusion}
\label{sec:discussion}
Leveraging the Poisson Unbiased Risk Estimator, a novel self-supervised conformal prediction method has been proposed. The method is useful for uncertainty quantification in Poisson imaging problems, a key class of imaging problems that were beyond the scope of existing self-supervised conformal prediction methodology \cite{everink2025}. Unlike standard split conformal prediction, the proposed method does not require any ground truth data for calibration, thus significantly extending the application scope of conformal uncertainty quantification. In addition to bypassing the efforts associated with acquiring reliable ground truth data, self-calibration also provides robustness to distribution shift, a key issue when deploying conformal methods in populations that might not be well represented by the training data available. The proposed approach is particularly powerful when combined with modern self-supervised image reconstruction techniques, trained directly from measurements without the need for ground truth data. The effectiveness of the proposed self-supervised conformal prediction method was assessed through Poisson image denoising and Poisson image deblurring experiments with the \texttt{DIV2K} dataset, where it delivered highly accurate conformal prediction regions and performs on par with fully supervised alternatives. 

Future work will seek to establish mathematical guarantees for the proposed method, with particular attention to the interplay between the number of considered imaging problems $M$, the dimensionality of $X$ and $Y$, the regularity of the estimator $\hat{x}$, and the number of projections used in the stochastic trace estimator within PURE. Furthermore, future work could also consider extensions of the proposed framework to other noise models from the exponential family. For example, it would be interesting to tackle Poisson-Gaussian noise models, which are commonly encountered in microscopy \cite{khademi2021self}. 

With regards to applications, uncertainty quantification is key when using restored signals in critical scientific and decision-making processes. For example, it is of utmost importance in epidemiology when monitoring an ongoing epidemic such as COVID-19~\cite{nash2022real}.
% Measurements consists in infection counts time series.
State-of-the-art models for viral epidemics describe the number of new infections at a given time $t$ as a Poisson random variable whose parameter depends on previous counts and on a key unknown quantity, the so-called \emph{reproduction number} $R_t$, which reflects the strength of the epidemic~\cite{Corimodel}. A major challenge in the estimation of $R_t$ during a ongoing pandemic is that available data are highly corrupted by administrative noise, making the estimation of $R_t$ an ill-conditioned inverse problem.
A ambitious perspective for future research is to extend the proposed self-supervised PURE-based conformal prediction framework to the estimation of $R_t$ by leveraging the recently proposed Autoregressive Poisson Unbiased Risk Estimator~\cite{pascal2024risk}. A key difficulty in that case will be to obtain accurate PURE estimates from short epidemiological time series data.
\bibliographystyle{IEEEtran}
\bibliography{references}
\end{document}